\documentclass[nohyperref]{article}

\usepackage{microtype}
\usepackage{graphicx}
\usepackage{subfigure}
\usepackage{booktabs} 

\usepackage{hyperref}

\usepackage[dynnworkshop]{icml2022}

\usepackage{amsmath}
\usepackage{amssymb}
\usepackage{mathtools}
\usepackage{amsthm}

\usepackage{enumitem}
\usepackage{amsfonts}
\usepackage{xspace}
\usepackage{multirow}
\usepackage{tablefootnote}
\allowdisplaybreaks

\usepackage[capitalize,noabbrev]{cleveref}

\theoremstyle{plain}
\newtheorem{theorem}{Theorem}[section]

\theoremstyle{definition}

\newtheorem{assumption}[theorem]{Assumption}
\theoremstyle{remark}

\newcommand{\x}{{\boldsymbol w}}

\newcommand{\cifarT}{CIFAR-10\xspace}
\newcommand{\cifarH}{CIFAR-100\xspace}

\newcommand{\fedhen}{FedHeN\xspace}
\newcommand{\decfl}{Decouple\xspace}
\newcommand{\hetfl}{NoSide\xspace}

\usepackage[textsize=tiny]{todonotes}

\icmltitlerunning{\fedhen: Federated Learning in Heterogeneous Networks}

\begin{document}

\twocolumn[
\icmltitle{\fedhen: Federated Learning in Heterogeneous Networks}



\icmlsetsymbol{equal}{*}

\begin{icmlauthorlist}
\icmlauthor{Durmus Alp Emre Acar}{BU}
\icmlauthor{Venkatesh Saligrama}{BU}
\end{icmlauthorlist}

\icmlaffiliation{BU}{Boston University, Boston, MA}

\icmlcorrespondingauthor{Durmus Alp Emre Acar}{alpacar@bu.edu}
\icmlcorrespondingauthor{Venkatesh Saligrama}{srv@bu.edu}

\icmlkeywords{Machine Learning, ICML}

\vskip 0.3in
]



\printAffiliationsAndNotice{}  

\begin{abstract}
We propose a novel training recipe for federated learning with heterogeneous networks where each device can have different architectures. We introduce training with a side objective to the devices of higher complexities to jointly train different architectures in a federated setting. We empirically show that our approach improves the performance of different architectures and leads to high communication savings compared to the state-of-the-art methods.

\end{abstract}

\section{Introduction}
\citet{mcmahan2017communication} propose Federated Learning (FL) as a new distributed optimization problem where a server gradually trains a global model on datapoints from many devices. Due to privacy concerns, it is not allowed to transfer datapoints. Instead, FL transmits local and global models between devices and the server. Since transmitting models is a costly operation \cite{halgamuge2009estimation}, FL aims to find a global model with as less number of communication as possible.

{\it Failure of traditional FL in heterogeneous devices.} In standard FL, the server trains one model for all devices. A more practical problem is to allow devices to have different architectures based on device resources. For instance, consider a FL setting with many cellphone users where we want to distributively train a model on all user data \cite{mcmahan_ramage_scientists_2017}. Some users might have the latest release of a cellphone whereas the rest use old versions. The users with the latest releases would prefer a different, potentially more complex, model than the user with old cellphones. Conventional FL as in \citet{mcmahan2017communication} fails in this case because devices have different model architectures.

In this work, we investigate the above practical FL problem where we allow devices to have different architectures based on their capacities. We propose \fedhen that modifies device training by introducing a novel side objective to the devices with complex models. The side objective allows \fedhen to jointly train complex and simple architectures.

We test our method in real-world datasets of CIFAR10 and CIFAR100 and compare it to a naive baseline and the current state-of-the-art method. We show that \fedhen achieves significant communication savings as well as better performance compared to the competitors.

{\bf Related Work.}

{\it FL.}
FedAvg \cite{mcmahan2017communication} is proposed as an extension of decentralized SGD \cite{parallelesgd}. The convergence of FedAvg depends on device data distribution \cite{Li2020On}. Different modifications are proposed to improve convergence guarantees such as FedProx \cite{li2018federated}, SCAFFOLD \cite{karimireddy2019scaffold}, FedDyn \cite{acar2021federated}. All of these methods average parameters of device models in the server. This is not allowed heterogeneous network setting since the architectures are not the same. Differently, \fedhen targets FL with heterogeneous architectures.

HeteroFL \cite{diao2021heterofl} introduces FL with heterogeneous networks problem. HeteroFL considers a setting where simple architecture is mapped to a subset of the complex architecture. The server constructs new models by averaging model weights of all devices based on the mapping. Different from HeteroFL, \fedhen introduces novel side objectives for complex device training.

{\it Early Exit.} 
Due to their size, big DNNs  consume more energy and they are slow to operate. \citet{q1, q2} propose to quantize/binarize the weights of DNNs to improve memory and computation costs. \citet{bolukbasi2017adaptive} propose to train a big DNN so that the network adaptively early exits in 'easy' examples to decrease inference costs. \citet{kaya2019shallow} propose to add a side objective on DNN to prevent 'overthinking' of DNNs for easy examples. Different from these works, we are interested in FL. \fedhen introduces side objectives to jointly train models in FL with different architectures.

{\bf Our Contributions.}
\begin{itemize}[leftmargin=10pt,nosep]
    \item We present \fedhen to jointly train a server model with different architectures based on device capacities.
    \item We empirically show that \fedhen achieves significant communication savings compared to the baselines.
\end{itemize}

\section{Method} \label{sec.method}
\begin{algorithm}[t]
    \caption{FL in Heterogeneous Networks - \fedhen}
    \label{alg.fedhen}
\begin{algorithmic}[1]
    \STATE {\bfseries Input:} $T$, $E$, $\eta$, N, $\{\mathcal{D}_i\}_{i\in [N]}$, initial models $\x_s^1,\x_c^1$,
    \FOR{$t=1\ldots T$}
        \STATE Randomly sample active devices, $\mathcal{Z}\subset[N]$,
        \STATE Divide $\mathcal{Z}$ into simple and complex devices, $\mathcal{Z}_s, \mathcal{Z}_c$ 
        \STATE {\color{cyan} \# Client Optimization}
        \FOR{$i \in \mathcal{Z}_s$}
            \STATE Receive the server simple model, $\x_s^t$,
            \STATE $\x_{s,i}^{t+1}=\text{ClientTraining}\left(\x_s^t, \mathcal{D}_i, E, \eta\right)$
            \STATE Transmit $\x_{s,i}^{t+1}$ back to the server.
        \ENDFOR
        \FOR{$j \in \mathcal{Z}_c$}
            \STATE Receive the server complex model, $\x_c^t$,
            \STATE $\x_{c,j}^{t+1}=\text{ClientTrainingSideObj}\left(\x_c^t, \mathcal{D}_j, E, \eta\right)$
            \STATE Transmit $\x_{c,j}^{t+1}$ back to the server.
        \ENDFOR
        \STATE {\color{cyan} \# Server Optimization}
        \STATE Set $\x_s^{t+1}$ using weights from all active devices,
        \STATE $\x_s^{t+1}=\frac{1}{|\mathcal{Z}|}\left(\sum_{i\in\mathcal{Z}_s}\x_{s,i}^{t+1}+\sum_{j\in\mathcal{Z}_c}\left[\x_{c,j}^{t+1}\right]_{\mathcal{M}}\right)$
        \STATE Set $\x_{c}^{t+1}$'s sub-net as the updated simple model,
        \STATE $\left[\x_{c}^{t+1}\right]_{\mathcal{M}} = \x_s^{t+1}$
        \STATE Set rest of the $\x_{c}^{t+1}$ using complex active devices,
        \STATE $\left[\x_{c}^{t+1}\right]_{\mathcal{M}^\prime} = \frac{1}{|\mathcal{Z}|_c}\sum_{j\in\mathcal{Z}_c}\left[\x_{c,j}^{t+1}\right]_{\mathcal{M}^\prime}$
    \ENDFOR
\end{algorithmic}
\end{algorithm}

\begin{algorithm}[ht]
    \caption{\fedhen Device Optimizations}
    \label{alg.fedhen.dev}
\begin{algorithmic}[1]
    \STATE {\bf function} $\text{ClientTraining}\left(\x_s, \mathcal{D}_i, E, \eta\right)$:
        \STATE\quad Start from $\x_i{=}\x_s$, train $E$ epochs,
        \STATE\quad {\bf for }$E$ epochs, batch $B\subset\mathcal{D}_i$ {\bf  do}
        \STATE\quad\quad Compute batch gradient, $\hat{\nabla}f_i(\x_i)$,
        \STATE\quad\quad Update, $\x_i\leftarrow \x_i-\eta\hat{\nabla}f_i(\x_i)$,
        \STATE\quad Return trained model $\x_i$
    \STATE {\bf end function}
    
    \STATE {\bf function} $\text{ClientTrainingSideObj}\left(\x_c, \mathcal{D}_j, E, \eta\right)$:
        \STATE\quad Start from $\x_j{=}\x_c$, train $E$ epochs with side obj.,
        \STATE\quad {\bf for }$E$ epochs, batch $B\subset\mathcal{D}_j$ {\bf  do}
        \STATE\quad\quad Compute batch gradient along with side obj., \STATE\quad\quad $\hat{\nabla}f_j(\x_j)$, $\hat{\nabla}f_j\left(\left[\x_j\right]_{\mathcal{M}}\right)$,
        \STATE\quad\quad Update, $\x_j{\leftarrow} \x_j{-}\eta\left(\hat{\nabla}f_j(\x_j){+}\hat{\nabla}f_j\left(\left[\x_j\right]_{\mathcal{M}}\right)\right)$,
        \STATE\quad Return trained model $\x_j$
    \STATE {\bf end function}
\end{algorithmic}
\end{algorithm}

FL setting consists of one server node and $N$ device nodes. Each device $i$ has a different dataset $\mathcal{D}_i$. Let $f_i:\mathcal{W}\rightarrow \mathcal{R}$ be the loss of using a model on device $i$'s dataset. FL solves,
\begin{align}\nonumber
\min_{\x \in \mathcal{W}} \frac{1}{N} \sum_{i \in [N]} f_i\left(\x\right)
\end{align}
where $\x$ is the NN parameters. 

Different from the conventional FL, we are interested in having different architectures in devices. For simplicity, consider a setting where we have a simple architecture, $\x_s\in \mathcal{W}_s$, and a complex architecture $\x_c \in \mathcal{W}_c$. Let $\mathcal{S}\subset [N]$ and $\mathcal{C}=[N]-S$ be the devices that have simple and complex architectures respectively. We reformulate our problem as,
\begin{align}\label{obj.fl}
&\min_{\substack{\x_s \in \mathcal{W}_s\\ \x_c \in \mathcal{W}_c}}  \frac{1}{|\mathcal{S}|} \sum_{i \in S} f_i\left(\x_s\right)+\frac{1}{|\mathcal{C}|} \sum_{j \in C} f_j\left(\x_c\right) \nonumber\\
&\text{such that  }\mathcal{R}(\x_s,\x_c)=0
\end{align}
where $\mathcal{R}(\x_s,\x_c)$ captures the relationship between simple and complex architectures.

We note that Eq. \ref{obj.fl} is an ERM objective. If there is no condition (no $\mathcal{R}(\x_s,\x_c)=0$ constraint), one could minimize the objective by separating complex and simple losses. However, we are not interested in training data, we would like to train models that perform well on test data. Hence, we introduce $\mathcal{R}(\x_s,\x_c)$ as a way of regularizing the models.

To relate simple and complex architectures, we assume,
\begin{assumption}\label{as.1}
Simple architecture is a sub-network of the complex architecture. There exists a set of indices, $\mathcal{M}$, of complex architecture such that $\mathcal{W}_s=\{\left[\x_c\right]_{\mathcal{M}} | \x_c \in \mathcal{W}_c\}$ where $\left[\x_c\right]_{\mathcal{M}}$ selects the weights of $\x_c$ based on the index set $\mathcal{M}$.
\end{assumption}

We encourage weight sharing between simple architecture and the corresponding sub-network of the complex architecture as in Assumption \ref{as.1}. We let $\mathcal{R}(\x_s,\x_c)=\|\x_s-\left[\x_c\right]_{\mathcal{M}}\|^2$.

To further regularize models, we add a side objective to the complex device training. Complex devices minimize their losses along with the corresponding sub-network of simple architecture as,
\begin{align}\label{obj.fl.side}
&\min_{\substack{\x_s \in \mathcal{W}_s\\ \x_c \in \mathcal{W}_c}}  \frac{1}{|\mathcal{S}|} {\sum_{i \in S}} {f_i\left(\x_s\right)}{+}\frac{1}{|\mathcal{C}|} \left({\sum_{j \in C}} f_j\left(\x_c\right){+}f_j\left(\left[\x_c\right]_{\mathcal{M}}\right)\right) \nonumber\\
&\text{such that  }\mathcal{R}(\x_s,\x_c)=0
\end{align}

\begin{table}[t]
    \caption{IID split, $50$ simple and $50$ complex devices, 10\% participation rate. The number of communication rounds required to achieve the target test performance for different methods. The gain in using \fedhen compared to best baseline method is given.}
    \label{tab:res.iid}
    \resizebox{.49\textwidth}{!}{
    \begin{tabular}{| c | c || c || c | c || c |}
    \hline
    Dataset & Accuracy & \fedhen & \decfl & \hetfl &  Gain\\\hline\hline
    
    \multicolumn{6}{|c|}{\it Simple Model}\\\cline{1-6}
    \multirow{2}{*}{\cifarT}
    &84.4 &   289  &   943  &   805  & {\bf 2.8$\times$}\\
    &83.4 &   249  &   731  &   669  & {\bf 2.7$\times$}\\\cline{1-6}
    \multirow{2}{*}{\cifarH}
    &46.4 &   296  &   864  &   984  & {\bf 2.9$\times$}\\
    &45.4 &   250  &   588  &   807  & {\bf 2.4$\times$}\\\cline{1-6}
    
    \multicolumn{6}{|c|}{\it Complex Model}\\\cline{1-6}
    \multirow{2}{*}{\cifarT}
    &88.5 &   649  &   991  &   941  & {\bf 1.4$\times$}\\
    &87.5 &   456  &   739  &   669  & {\bf 1.5$\times$}\\\cline{1-6}
    \multirow{2}{*}{\cifarH}
    &46.8 &   468  &   963  &   614  & {\bf 1.3$\times$}\\
    &45.8 &   376  &   752  &   472  & {\bf 1.3$\times$}\\\cline{1-6}    
    \hline
    
    \end{tabular}
    }
\vspace{-14pt}
\end{table}

We would like highlight some properties of Eq. \ref{obj.fl.side},
\begin{itemize}[leftmargin=10pt,nosep]
    \item Simple architecture is trained on all datapoints instead of on the datapoints only from the simple devices. Hence, the generalization of simple architecture is improved.
    \item $\mathcal{R}(\x_s,\x_c)$ correlates simple and complex architecture. A better simple model leads to a better complex model through $\mathcal{R}(\x_s,\x_c)=0$ condition.
\end{itemize}

{\bf \fedhen Algorithm.} \fedhen solves Eq. \ref{obj.fl.side} with the steps summarized in Algorithm \ref{alg.fedhen}. 

In each round, a random subset of devices become active, $\mathcal{Z}$. We divide set $\mathcal{Z}$ into simple active and complex active device sets as $\mathcal{Z}_s$ and $\mathcal{Z}_c$ respectively.

Simple active devices receive the server simple model, $\x_s^t$. We compute a local model $\x_{s,i}^{t+1}$ by starting from $\x_s^t$ and training it for $E$ epochs on local dataset $\mathcal{D}_i$ displayed as 'ClientTraining' method (Alg. \ref{alg.fedhen.dev}). The trained model is transmitted back to the server.

Complex active devices receive the server complex model, $\x_c^t$. We train a local model starting from $\x_c^t$ and training it for $E$ epochs using their local dataset, $\mathcal{D}_j$ with gradients of the complex and the simple model shown as 'ClientTrainingSideObj' method (Alg. \ref{alg.fedhen.dev}). Namely, we update the model with summation of batch gradient of complex model, $\hat{\nabla}f_j(\x_j)$, and batch gradient of the corresponding sub-network (simple) model, $\hat{\nabla}f_j\left(\left[\x_j\right]_{\mathcal{M}}\right)$. The trained model is transmitted back to the server.

The server collects models from participating devices. The server simple model is constructed by averaging weights from all active devices, .i.e the simple devices $\{\x_{s,i}^{t+1}\}_{i\in\mathcal{Z}_s}$ as well as the common sub-net of the complex devices $\left\{\left[\x_{c,j}^{t+1}\right]_{\mathcal{M}}\right\}_{j\in\mathcal{Z}_c}$, ln. 18 in Alg. \ref{alg.fedhen}. The common sub-network of the server complex model is set equal to the constructed server simple model, ln. 20 in Alg. \ref{alg.fedhen}. Finally, the rest of the server complex model is constructed by averaging the corresponding weights of the active complex models, .i.e $\left\{\left[\x_{c,j}^{t+1}\right]_{\mathcal{M}^\prime}\right\}_{j\in\mathcal{Z}_c}$ where $\mathcal{M}^\prime$ corresponds to the sub-network that is not common with the simple architecture, ln. 22 in Alg. \ref{alg.fedhen}.

This completes one round of training of \fedhen. We iterate the same process for $T$ communication rounds.

{\it Cost of side objective. } In passing, we note that side objective adds minimal  cost to the complex devices. Firstly, it is a light weight operation. Complex devices calculate gradients of the full model, $\x_c$. Calculating the gradient with respect to the simple model, $\x_s$, requires less computation. Secondly, the main energy consumption occurs during transmission of models \cite{halgamuge2009estimation}.

\section{Experiments}
\begin{table}[t]
    \caption{Non-IID split, $50$ simple and $50$ complex devices, 10\% participation rate. The number of communication rounds required to achieve the target test performance for different methods. The gain in using \fedhen compared to best baseline method is given.}
    \label{tab:res.non-iid}
    \resizebox{.49\textwidth}{!}{
    \begin{tabular}{| c | c || c || c | c || c |}
    \hline
    Dataset & Accuracy & \fedhen & \decfl & \hetfl &  Gain\\\hline\hline

    \multicolumn{6}{|c|}{\it Simple Model }\\\cline{1-6}
    \multirow{2}{*}{\cifarT}
    &79.4 &   295  &   986  &   810  & {\bf 2.7$\times$}\\
    &78.4 &   256  &   816  &   676  & {\bf 2.6$\times$}\\\cline{1-6}
    \multirow{2}{*}{\cifarH}
    &43.8 &   278  &   978  &   914  & {\bf 3.3$\times$}\\
    &42.8 &   239  &   813  &   762  & {\bf 3.2$\times$}\\\cline{1-6}
    
    \multicolumn{6}{|c|}{\it Complex Model }\\\cline{1-6}
    \multirow{2}{*}{\cifarT}
    &84.2 &   596  &  1000  &   857  & {\bf 1.4$\times$}\\
    &83.2 &   519  &   887  &   751  & {\bf 1.4$\times$}\\\cline{1-6}
    \multirow{2}{*}{\cifarH}
    &44.8 &   450  &   997  &   498  & {\bf 1.1$\times$}\\
    &43.8 &   372  &   754  &   456  & {\bf 1.2$\times$}\\\cline{1-6}   
    \hline
    \end{tabular}
    }
\vspace{-13pt}
\end{table}

In this section, we compare \fedhen method to baselines in real-world dataset settings. We refer to Appendix \ref{ap.exp} for a description of the hyperparameters.

\begin{figure*}[t]
\centering\hfill
\subfigure{\includegraphics[width=.33\linewidth]{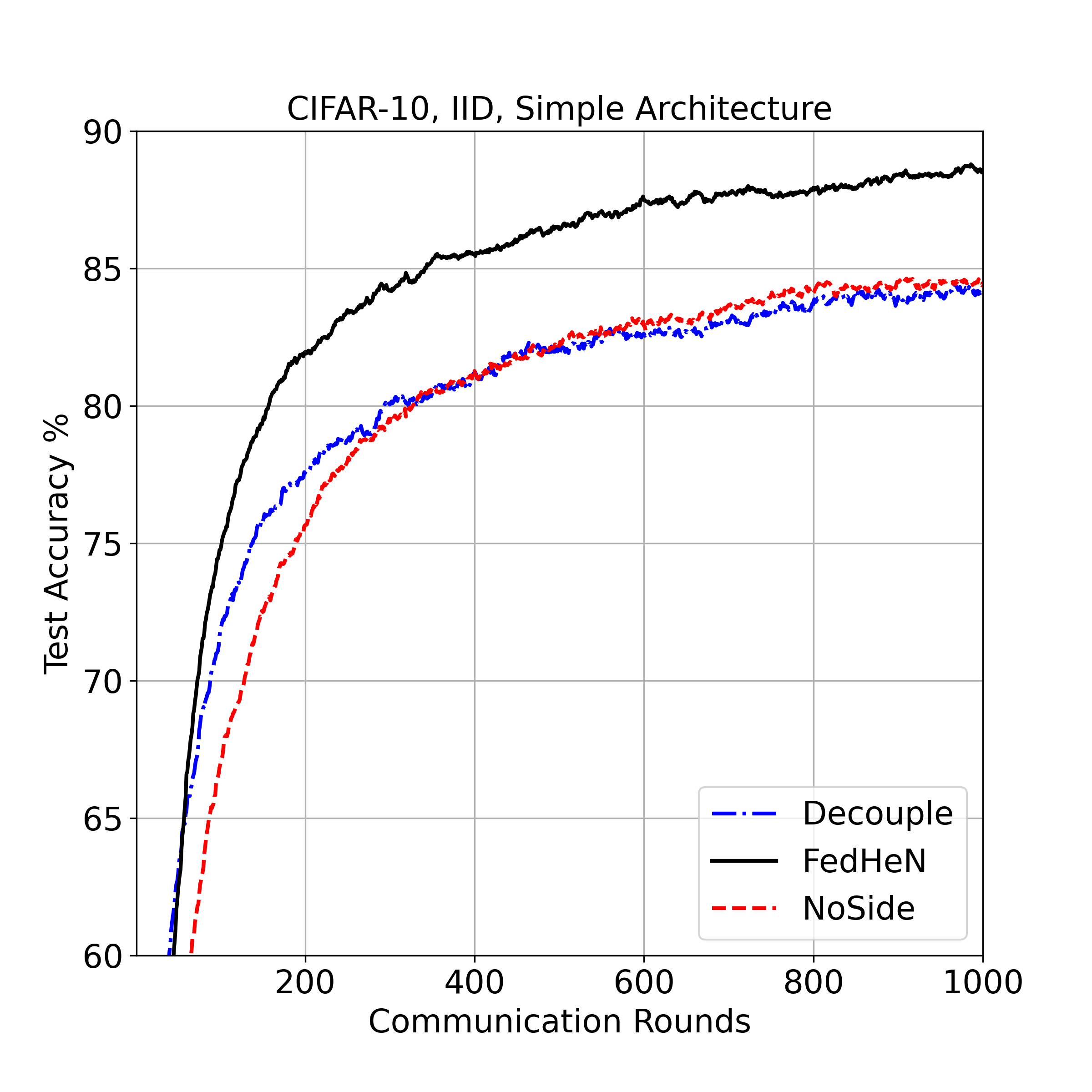}}\hfill\hfill
\subfigure{\includegraphics[width=.33\linewidth]{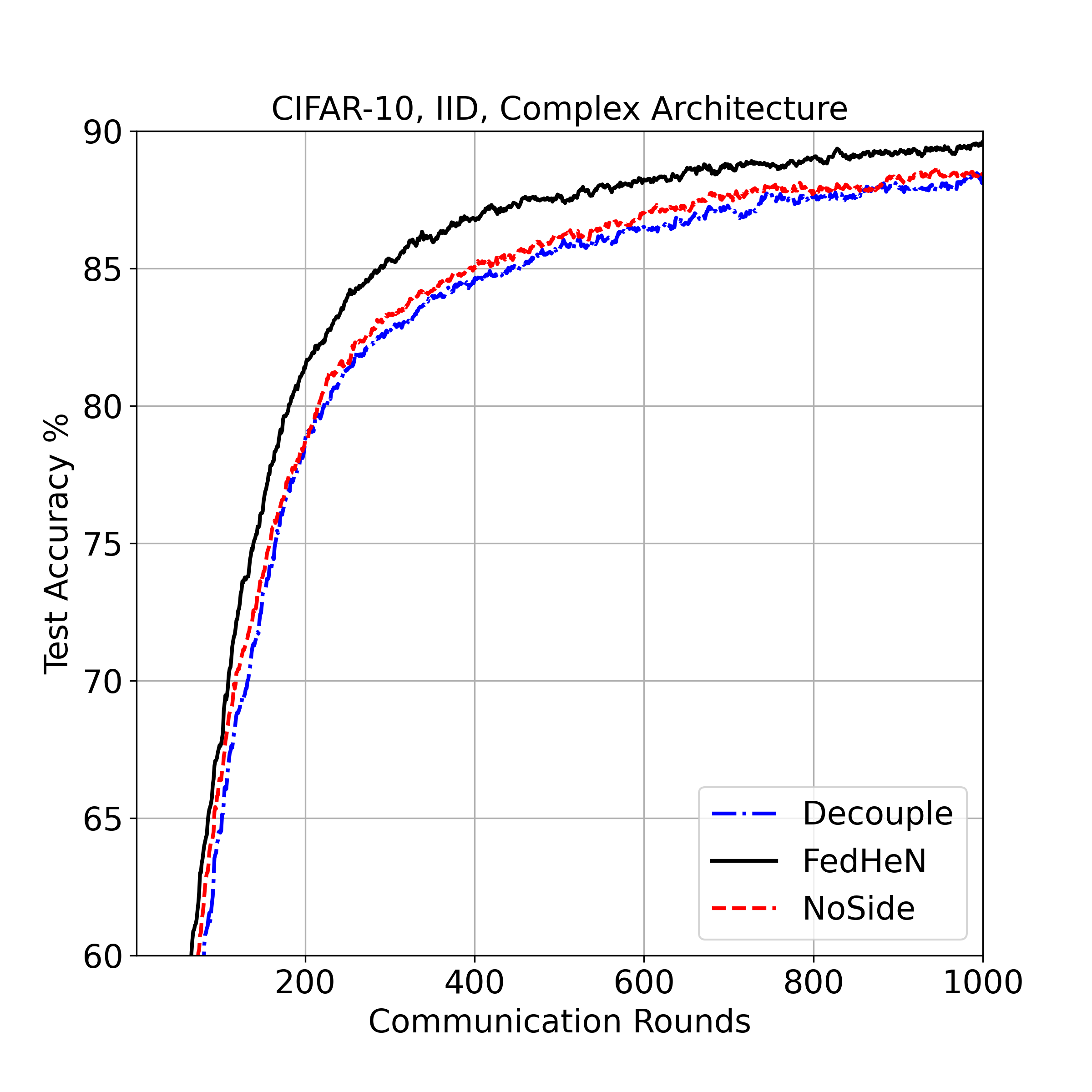}}\hfill
\vspace{-15pt}
\caption{ Test accuracy vs. communication rounds on \cifarT IID split. {\bf Left}: Simple,  {\bf Right}: Complex.}
\label{fig:conv.cifar10_iid}
\vspace{-15pt}
\end{figure*}

{\bf FL dataset.} We test our method using \cifarT and \cifarT \cite{krizhevsky2009learning}. We split the datasets into $100$ clients and randomly activate $10$ clients in each round. We consider both IID and non-IID splits in our experiments. IID split is constructed by randomly partitioning data into clients. Non-IID split is constructed using a Dirichlet prior on the labels as in \citet{yurochkin2019bayesian}.

We assume the first 50 devices have simple architecture and the last 50 devices have complex architecture in all experiments. 

{\bf Models.} We use PreActResNet18\footnote{BatchNorm layers store data statistics which results in privacy leakage. We use GroupNorm layers \cite{wu2018group} instead in all ResNet models.} \cite{he2016identity} for the complex architecture. PreActResNet18 has 4 residual blocks and total of 11.1M parameters. 

If we centralize all client datapoints, the complex architecture gets 93\% and 73.5\% performance for \cifarT and \cifarH datasets respectively. If we centralize half of the datapoints as in 50 devices, the complex architecture gets 90.5\% and 62.6\% performance for \cifarT and \cifarH datasets respectively.

As a simple architecture, we consider the first 2 residual blocks of PreActResNet18. Then, we add a mix pooling layer \cite{lee2016generalizing} that learns a weighted combination of avg pooling and max pooling layers as in \citet{kaya2019shallow}. The simple architecture has overall 0.7M parameters. 

If we centralize all client datapoints, the simple architecture gets 86\% and 63.2\% performance for \cifarT and \cifarH datasets respectively. If we centralize half of the datapoints as in 50 devices, simple model gets 84.5\%, 55.7\% performance for \cifarT and \cifarH respectively.

{\bf Methods.} We compare \fedhen to two baselines as,
\begin{itemize}[leftmargin=10pt,nosep]
    \item {\it Naive \decfl}. \decfl minimizes Eq. \ref{obj.fl} without any $\mathcal{R}(\x_s,\x_c)=0$ constraint. It decouples complex and simple device training. It separately trains a complex and a simple model using FedAvg. \decfl is summarized in Algorithm \ref{alg.decouple}.
    \item {\it \hetfl\footnote{HeteroFL is proposed in a setting where simple model is obtained by shrinking CNN channels of the complex model different from our setting. Except from the simple model definition, HeteroFL uses the same $\mathcal{R}(\x_s,\x_c)$ as \fedhen and it does not add side objective. We name HeteroFL in our setting as NoSide.} \cite{diao2021heterofl}}. \hetfl is motivated from HeteroFL \cite{diao2021heterofl}. It minimizes Eq. \ref{obj.fl} with the same $\mathcal{R}(\x_s,\x_c)$ as \fedhen. The key difference is that it does not use side objective in the complex architecture training. \hetfl is summarized in Algorithm \ref{alg.noside}.
\end{itemize}

{\bf Evaluation Metric.} We fix a target test accuracy for server simple and server complex models. We compare the number of communications rounds to achieve the target test accuracy for all methods.

{\bf Results.} Table \ref{tab:res.iid} \& \ref{tab:res.non-iid} show the number of communication rounds to get target accuracies in all methods. We highlight the gain of using \fedhen compared to the best competitor in the last column. We present convergence curves vs. communication rounds in Figure \ref{fig:conv.cifar10_iid}, \ref{fig:conv.cifar10_non_iid} \& \ref{fig:conv.cifar100} (Appendix \ref{ap.exp}).

{\it \fedhen leads to significant communication savings.}  We observe that \fedhen trains better models uniformly in all the experiments shown in the gain columns of Table \ref{tab:res.iid} \& \ref{tab:res.non-iid}. For instance, the same simple performance of 43.8\% is obtained using $3.3\times$ less communication with \fedhen in \cifarH non-IID setting. The communication savings ranges from $1.1\times$ to $3.3\times$ in our experiments.

{\it \decfl vs \hetfl. } \decfl is a naive algorithm. However, it performs close to \hetfl algorithm in \cifarT IID and non-IID settings as shown in Figure \ref{fig:conv.cifar10_iid} \& \ref{fig:conv.cifar10_non_iid}. For instance, the same test accuracy for complex model is achieved in 991 and 941 rounds for \decfl and \hetfl models in \cifarT IID setting as shown in Table \ref{tab:res.iid}. 

{\it Simple model in \fedhen achieves similar to centralized accuracy.} \fedhen achieves better simple performance because simple architecture is trained on all datapoints due to the side objective in complex devices. Moreover, weight sharing between complex and simple architecture improves simple model's generalization. For instance, \fedhen's simple model in \cifarT achieves 88.6\% test accuracy in IID split within 1000 communication rounds as shown in Figure \ref{fig:conv.cifar10_iid} which is higher than the centralized accuracy of simple model. Differently, \hetfl and \decfl gets worse simple performance compared to \fedhen and the centralized model as shown in Figure \ref{fig:conv.cifar10_iid}, \ref{fig:conv.cifar10_non_iid} \& \ref{fig:conv.cifar100}.

{\it Complex model in \fedhen achieves better performance compared to competitors.} Training with side objective improves complex model performance in \fedhen. For instance, in \cifarT IID setting, \fedhen's complex model achieves 89.6\% test performance within 1000 communication rounds which is $>$1\% better compared to the competitors. This is also reflected in the gain values. \fedhen leads to $1.5\times$ communication savings for complex model.

\section{Conclusion}
We consider FL with heterogeneous networks where devices use different architectures based on their capacities. Our method, \fedhen, modifies device objectives of the complex device models to jointly train with different architectures. \fedhen leads to high communication savings compared to the baseline methods. 

\clearpage
\section*{Acknowledgements}
This research was supported by the Army Research Office Grant W911NF2110246, the National Science Foundation grants CCF-2007350 and CCF-1955981, and the Hariri Data Science Faculty Fellowship Grants, and a gift from the ARM corporation. 
\bibliography{references.bib}
\bibliographystyle{icml2022}

\clearpage
\appendix
\onecolumn
\section{Appendix}
\begin{minipage}{0.495\textwidth}
\begin{algorithm}[H]
    \caption{Algorithm \decfl}
    \label{alg.decouple}
\begin{algorithmic}[1]
    \STATE {\bfseries Input:} $T$, $E$, $\eta$, initial models $\x_s^1,\x_c^1$,
    \FOR{$t=1\ldots T$}
        \STATE Randomly sample active devices, $\mathcal{Z}\subset[N]$,
        \STATE Divide $\mathcal{Z}$ into simple and complex devices, $\mathcal{Z}_s, \mathcal{Z}_c$ 
        \STATE {\color{cyan} \# Client Optimization}
        \FOR{$i \in \mathcal{Z}_s$}
            \STATE Receive the server simple model, $\x_s^t$,
            \STATE $\x_{s,i}^{t+1}=\text{ClientTraining}\left(\x_s^t, \mathcal{D}_i, E, \eta\right)$
            \STATE Transmit $\x_{s,i}^{t+1}$ back to the server.
        \ENDFOR
        \FOR{$j \in \mathcal{Z}_c$}
            \STATE Receive the server complex model, $\x_c^t$,
            \STATE $\x_{c,j}^{t+1}=\text{ClientTraining}\left(\x_c^t, \mathcal{D}_j, E, \eta\right)$
            \STATE Transmit $\x_{c,j}^{t+1}$ back to the server.
        \ENDFOR
        \STATE {\color{cyan} \# Server Optimization}
        \STATE Set $\x_s^{t+1}$ using weights from simple active devices,
        \STATE $\x_s^{t+1}=\frac{1}{|\mathcal{Z}_s|}\sum_{i\in\mathcal{Z}_s}\x_{s,i}^{t+1}$
        \STATE Set $\x_c^{t+1}$ using weights from complex active devices,
        \STATE $\x_c^{t+1}=\frac{1}{|\mathcal{Z}_c|}\sum_{i\in\mathcal{Z}_c}\x_{c,i}^{t+1}$
    \ENDFOR
\end{algorithmic}
\end{algorithm}
\end{minipage}
\hfill
\begin{minipage}{0.495\textwidth}
\begin{algorithm}[H]
    \caption{Algorithm \hetfl}
    \label{alg.noside}
\begin{algorithmic}[1]
    \STATE {\bfseries Input:} $T$, $E$, $\eta$, initial models $\x_s^1,\x_c^1$,
    \FOR{$t=1\ldots T$}
        \STATE Randomly sample active devices, $\mathcal{Z}\subset[N]$,
        \STATE Divide $\mathcal{Z}$ into simple and complex devices, $\mathcal{Z}_s, \mathcal{Z}_c$ 
        \STATE {\color{cyan} \# Client Optimization}
        \FOR{$i \in \mathcal{Z}_s$}
            \STATE Receive the server simple model, $\x_s^t$,
            \STATE $\x_{s,i}^{t+1}=\text{ClientTraining}\left(\x_s^t, \mathcal{D}_i, E, \eta\right)$
            \STATE Transmit $\x_{s,i}^{t+1}$ back to the server.
        \ENDFOR
        \FOR{$j \in \mathcal{Z}_c$}
            \STATE Receive the server complex model, $\x_c^t$,
            \STATE $\x_{c,j}^{t+1}=\text{ClientTraining}\left(\x_c^t, \mathcal{D}_j, E, \eta\right)$
            \STATE Transmit $\x_{c,j}^{t+1}$ back to the server.
        \ENDFOR
        \STATE {\color{cyan} \# Server Optimization}
        \STATE Set $\x_s^{t+1}$ using weights from all active devices,
        \STATE $\x_s^{t+1}=\frac{1}{|\mathcal{Z}|}\left(\sum_{i\in\mathcal{Z}_s}\x_{s,i}^{t+1}+\sum_{j\in\mathcal{Z}_c}\left[\x_{c,j}^{t+1}\right]_{\mathcal{M}}\right)$
        \STATE Set $\x_{c}^{t+1}$'s sub-net as the updated simple model,
        \STATE $\left[\x_{c}^{t+1}\right]_{\mathcal{M}} = \x_s^{t+1}$
        \STATE Set rest of the $\x_{c}^{t+1}$ using complex active devices,
        \STATE $\left[\x_{c}^{t+1}\right]_{\mathcal{M}^\prime} = \frac{1}{|\mathcal{Z}|_c}\sum_{j\in\mathcal{Z}_c}\left[\x_{c,j}^{t+1}\right]_{\mathcal{M}^\prime}$

    \ENDFOR
\end{algorithmic}
\end{algorithm}
\end{minipage}

\label{ap.exp}

{\bf \decfl Algorithm.} We present \decfl methods in Algorithm \ref{alg.decouple}. \decfl fully decouples simple and complex architecture training. We explain the method in detail below.

In each round, a random subset of devices become active, $\mathcal{Z}$. $\mathcal{Z}$ is then divided into simple active and complex active device sets as $\mathcal{Z}_s$ and $\mathcal{Z}_c$ respectively. Simple active devices receive the server simple model. A local model is trained using 'ClientTraining' method (Alg. \ref{alg.fedhen.dev}). The trained model is transmitted back to the server. Complex active devices follow a similar process where they receive the server complex model. Then, a local model is trained using 'ClientTraining' method. The trained model is transmitted back to the server.

The server simple model is constructed by averaging weights from all active simple devices,  $\{\x_{s,i}^{t+1}\}_{i\in\mathcal{Z}_s}$. The server complex model is constructed using the all active complex devices, $\left\{\x_{c,j}^{t+1}\right\}_{j\in\mathcal{Z}_c}$.

{\bf \hetfl Algorithm.} \hetfl method is presented in Algorithm \ref{alg.noside}. We explain the method in detail below.

In each round, a random subset of devices become active, $\mathcal{Z}$. We divide set $\mathcal{Z}$ into simple active and complex active device sets as $\mathcal{Z}_s$ and $\mathcal{Z}_c$ respectively. Simple and complex device training is the same as in \decfl method where each active device receive the current server model based on their capacity, then train a local model using 'ClientTraining' method and transmit the trained model back to the server. 

The server step is the same as in \fedhen method. The server simple model is constructed by averaging weights from all active devices, .i.e the simple devices $\{\x_{s,i}^{t+1}\}_{i\in\mathcal{Z}_s}$ as well as the common sub-net of the complex active devices $\left\{\left[\x_{c,j}^{t+1}\right]_{\mathcal{M}}\right\}_{j\in\mathcal{Z}_c}$. The common sub-architecture of the server complex model is set equal to the constructed server simple model. The rest of the server complex model is constructed by averaging the corresponding weights of the active complex models, .i.e $\left\{\left[\x_{c,j}^{t+1}\right]_{\mathcal{M}^\prime}\right\}_{j\in\mathcal{Z}_c}$.

{\bf Related Works.} We mention more dimensions of related work in this subsection.

{\it Personalized federated learning. } Personalized federated learning \cite{chen2018federated} extends meta learning \cite{thrun2012learning} into FL. The server trains a meta model and the devices customize the meta model using the local dataset. \citet{fallah2020personalized} propose to use FedAvg and MAML \cite{finn2017maml} meta adaptation. \citet{acar2021debiasing} use debiasing algorithms to improve convergence of FedAvg. Differently, \fedhen trains different complexity models for devices based on the capacities. This can be thought as customizing the device models based on the device resources. \fedhen can be improved with personalized federated learning methods.

{\it Client selection.} There are works that target to reduce communication costs through client selection \cite{zhang, fl_sel, fl_sel2}. A smart client selection further decreases the number of iterations and the communication costs compared to random activation. Different from client selection, \fedhen allows simple devices to participate federated learning which have lower communication footprint that complex device models. \fedhen can be adapted in to client selection algorithms.

{\bf Hyperparameters.} We train each method for $1000$ communication rounds. Each active device trains models for $E=5$ epochs with learning rate as $\eta=0.1$. We use SGD optimizer during training and clip gradients (at norm 10) to improve stability. If a device model fails in training, .i.e gets NaN weights, we ignore that device in server model construction only for that round. The methods are implemented using PyTorch library \cite{pytorch}.

\begin{figure}[t]
\centering

\subfigure[]{\label{fig:conv.cifar10.3}\includegraphics[width=.4\textwidth]{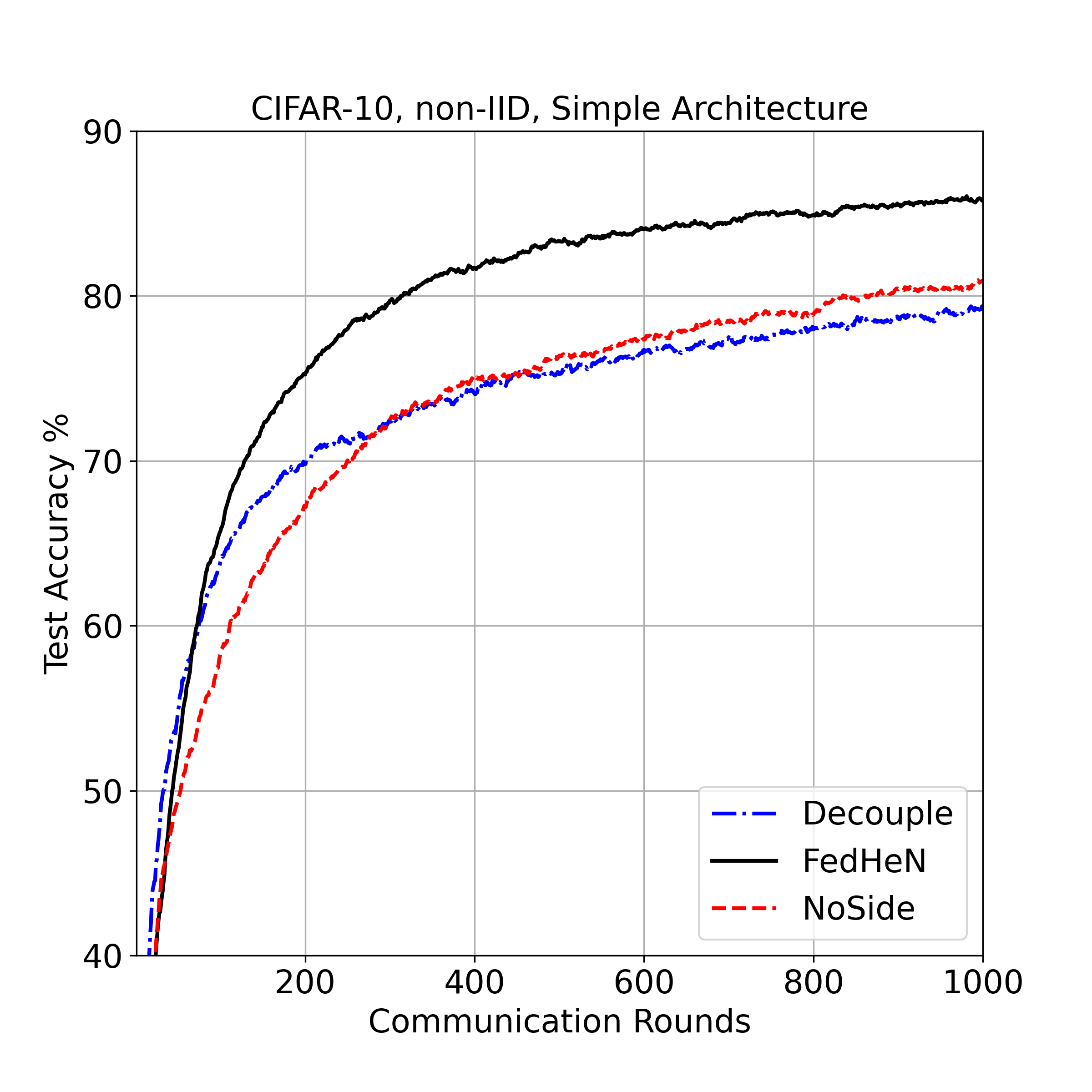}}\quad\quad
\subfigure[]{\label{fig:conv.cifar10.4}\includegraphics[width=.4\textwidth]{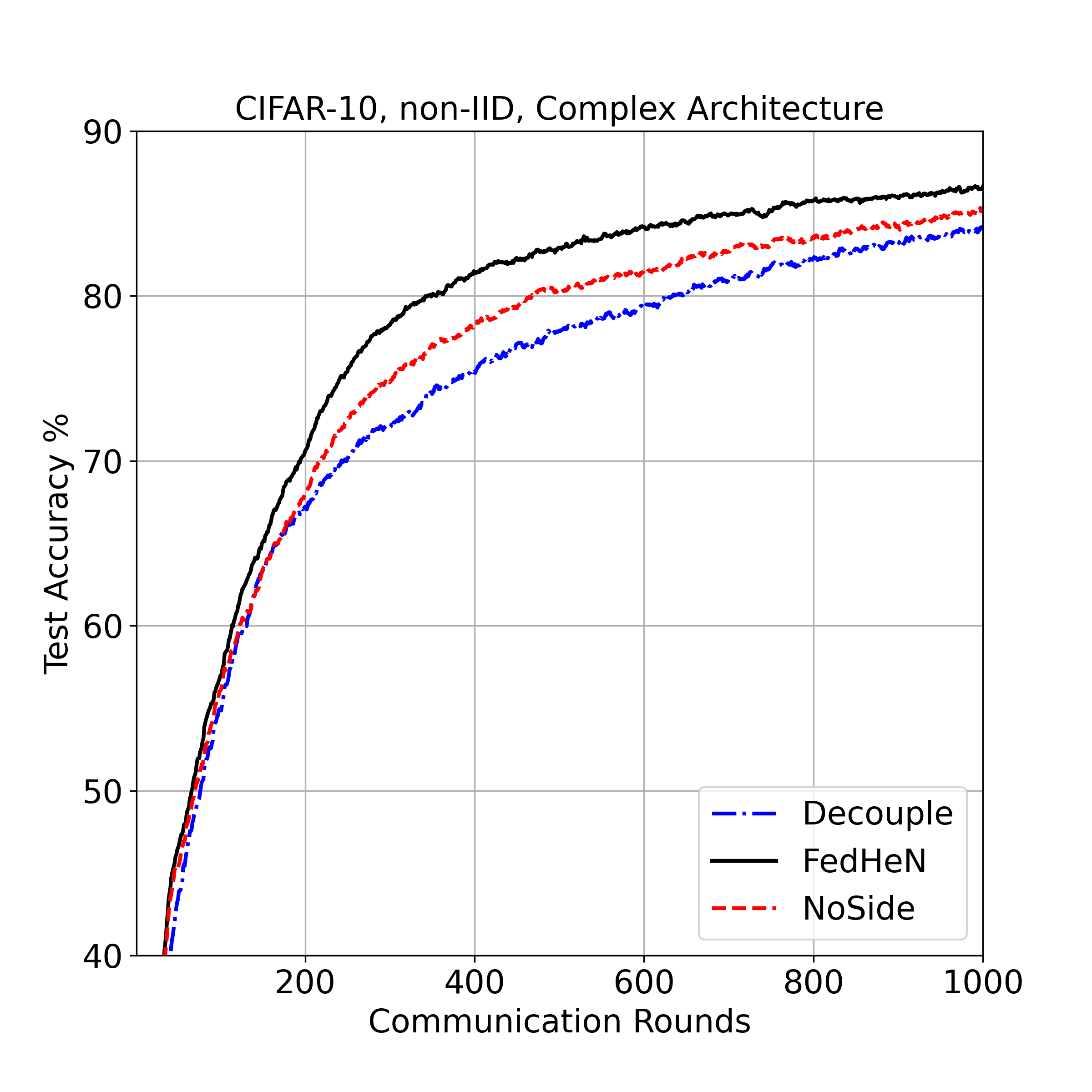}}
\caption{ Test accuracy vs. communication rounds on \cifarT non-IID split.  \ref{fig:conv.cifar10.3}: Simple, \ref{fig:conv.cifar10.4}: Complex.}
\label{fig:conv.cifar10_non_iid}
\end{figure}

\begin{figure}
\centering
\subfigure[]{\label{fig:conv.cifar100.1}\includegraphics[width=.4\textwidth]{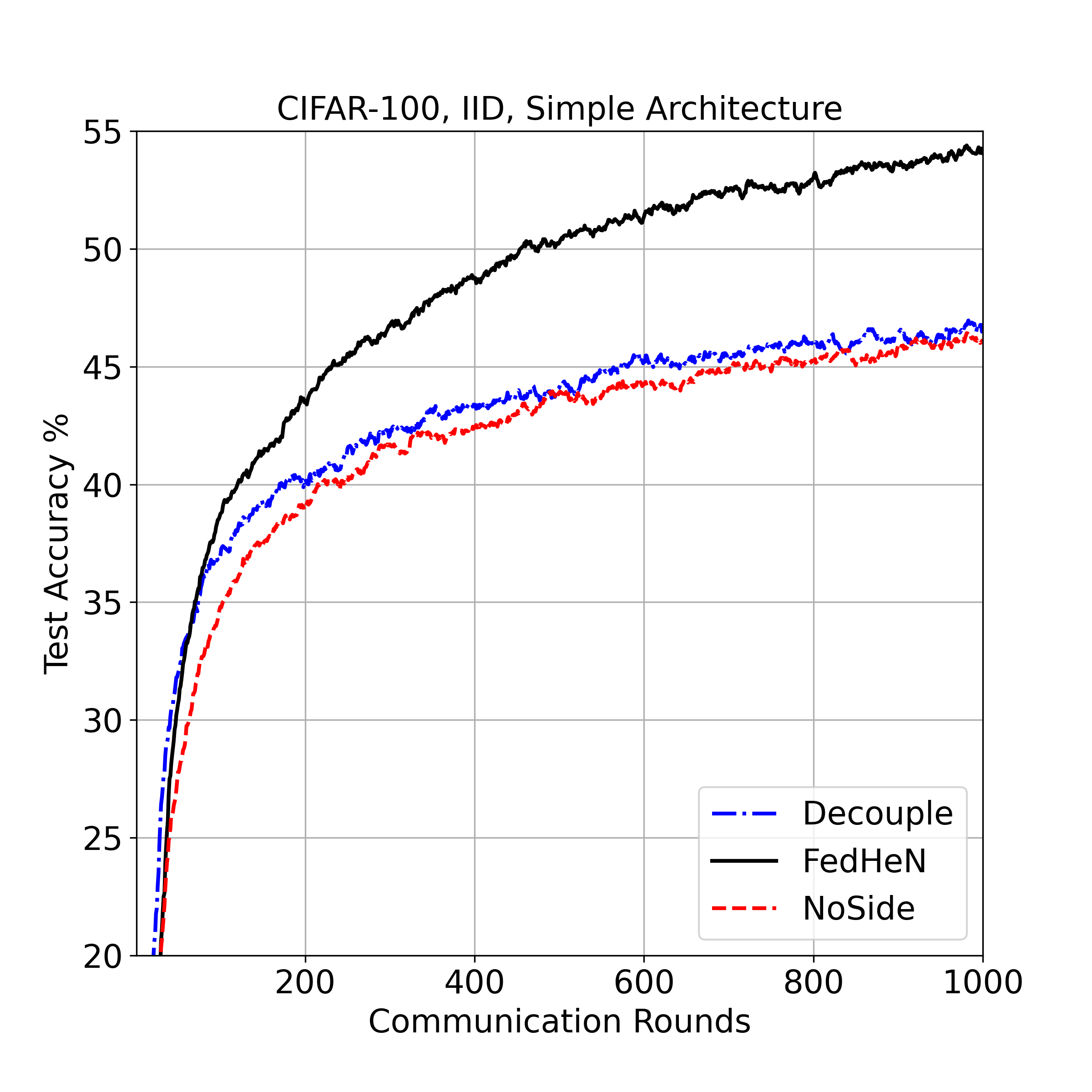}}\quad\quad
\subfigure[]{\label{fig:conv.cifar100.2}\includegraphics[width=.4\textwidth]{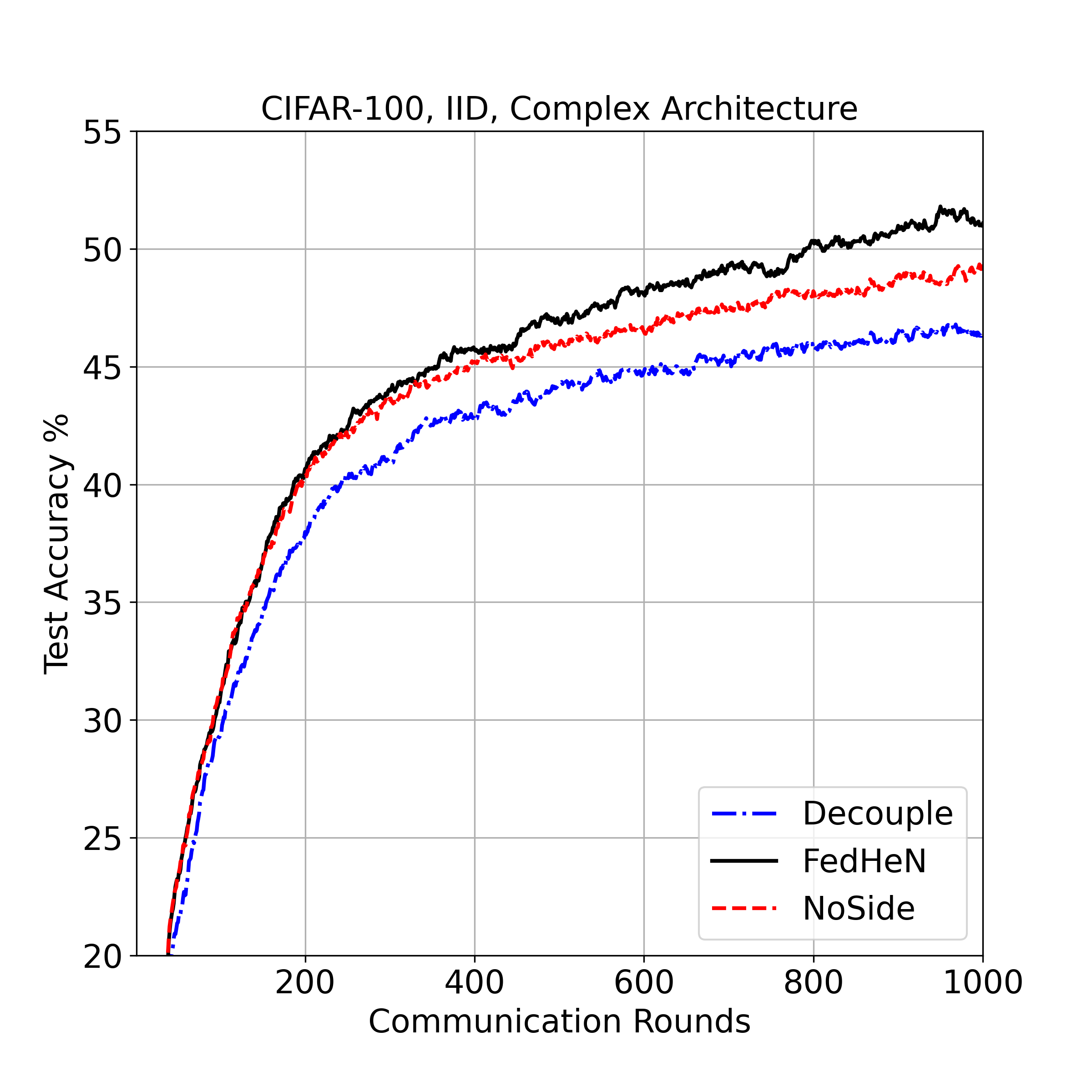}}

\subfigure[]{\label{fig:conv.cifar100.3}\includegraphics[width=.4\textwidth]{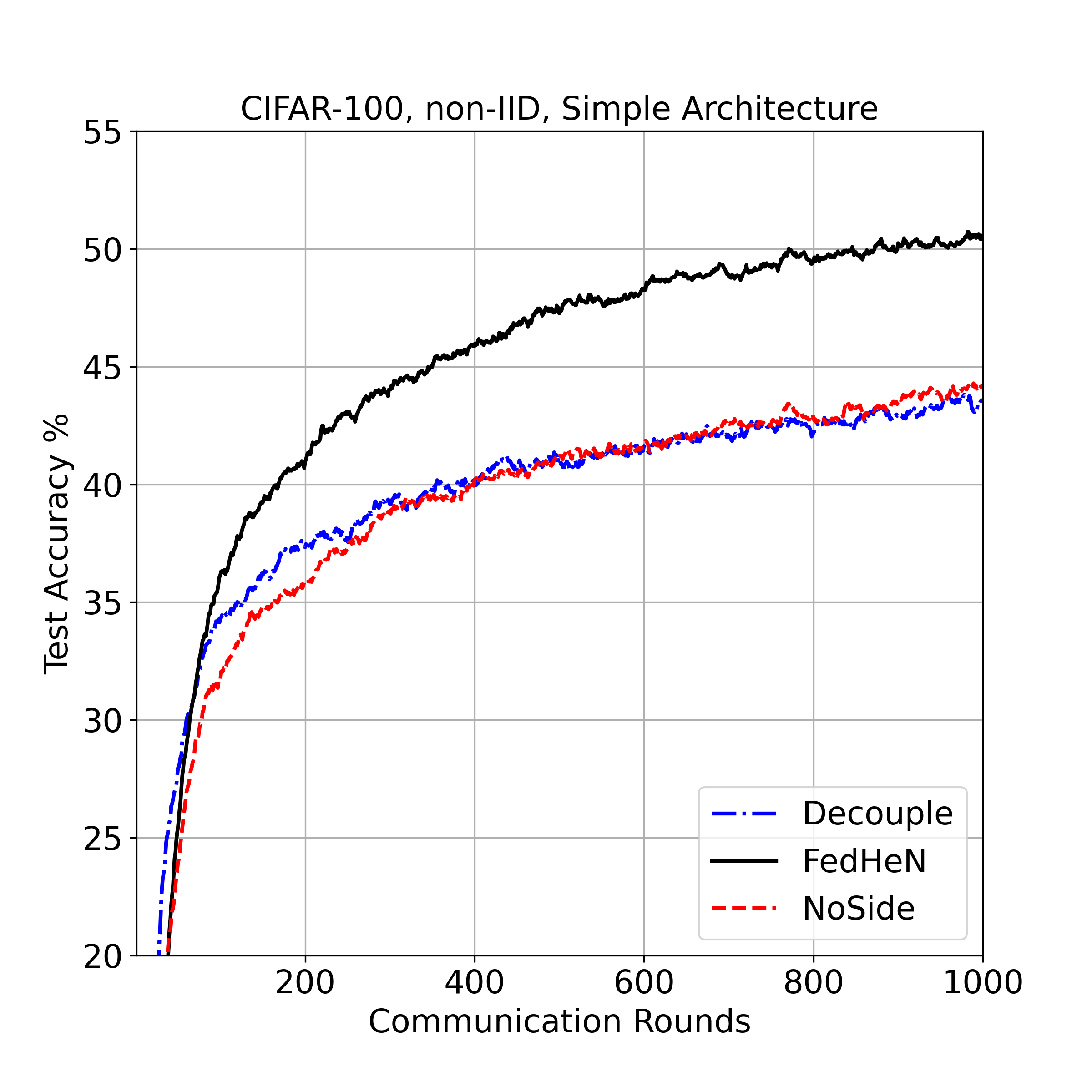}}\quad\quad
\subfigure[]{\label{fig:conv.cifar100.4}\includegraphics[width=.4\textwidth]{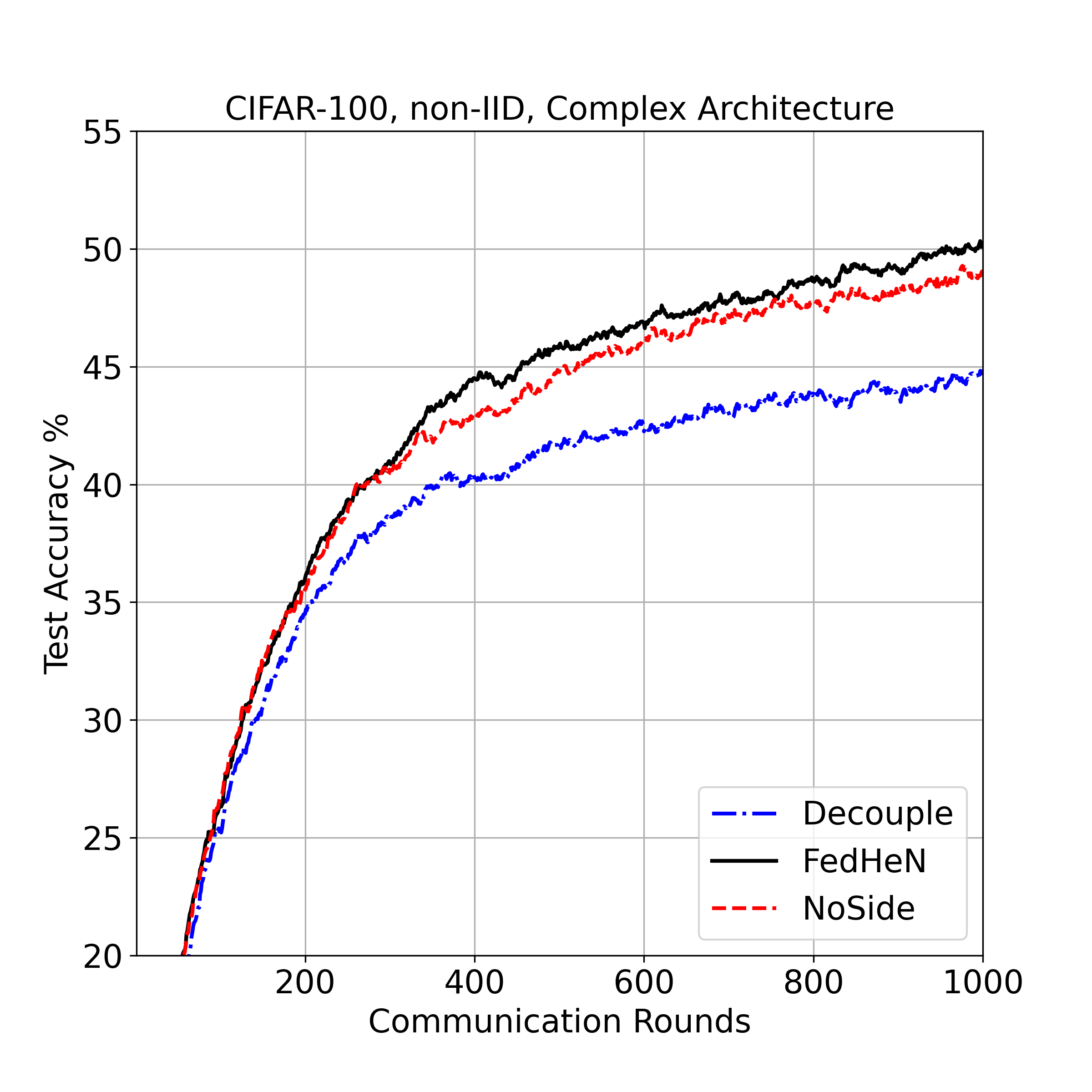}}
\caption{ Test accuracy vs. communication rounds on \cifarT. \ref{fig:conv.cifar100.1} \& \ref{fig:conv.cifar100.2}: IID split simple and complex. \ref{fig:conv.cifar100.3} \& \ref{fig:conv.cifar100.4}: non-IID split simple and complex.}
\label{fig:conv.cifar100}
\end{figure}

{\bf Figures and Algorithms.} \decfl and \hetfl are summarized in Algorithm \ref{alg.decouple} and \ref{alg.noside} respectively. The convergence curves of \fedhen, \decfl and \hetfl are presented in Figure \ref{fig:conv.cifar10_iid}, \ref{fig:conv.cifar10_non_iid} and \ref{fig:conv.cifar100}. 

{\bf Reporting Results.} Methods average only active devices to set server models. This introduces noise in convergence curves and communication gain calculations. We report/present model performances when we average all devices (all complex devices for server complex model and all simple devices for server simple model). We note that this is just for the reporting purposes and the training is performed as stated in Algorithm \ref{alg.fedhen}, \ref{alg.decouple} \& \ref{alg.noside}.

\end{document}